# DeepDiffusion: Unsupervised Learning of Retrieval-adapted Representations via Diffusion-based Ranking on Latent Feature Manifold

**Takahiko Furuya[1] and Ryutarou Ohbuchi[1]**

[1]Department of Computer Science and Engineering, University of Yamanashi, 4-3-11 Takeda, Kofu-shi, Yamanashi-ken, 400-8511, Japan

Corresponding author: Takahiko Furuya (e-mail: takahikof@ yamanashi.ac.jp).

This work was supported by JSPS KAKENHI (Grant No. 21K17763).

**ABSTRACT** Unsupervised learning of feature representations is a challenging yet important problem for analyzing a large collection of multimedia data that do not have semantic labels. Recently proposed neural network-based unsupervised learning approaches have succeeded in obtaining features appropriate for classification of multimedia data. However, unsupervised learning of feature representations adapted to content-based matching, comparison, or retrieval of multimedia data has not been explored well. To obtain such retrieval-adapted features, we introduce the idea of combining diffusion distance on a feature manifold with neural network-based unsupervised feature learning. This idea is realized as a novel algorithm called DeepDiffusion (DD). DD simultaneously optimizes two components, a feature embedding by a deep neural network and a distance metric that leverages diffusion on a latent feature manifold, together. DD relies on its loss function but not encoder architecture. It can thus be applied to diverse multimedia data types with their respective encoder architectures. Experimental evaluation using 3D shapes and 2D images demonstrates versatility as well as high accuracy of the DD algorithm. Code is available at https://github.com/takahikof/DeepDiffusion

## I. INTRODUCTION

Technology for content-based multimedia information retrieval is essential to manage a variety of data, such as text documents, two-dimensional (2D) images, and three-dimensional (3D) shapes. Over the past decade, accuracy of multimedia information retrieval has improved drastically due in large part to the development of distance metric learning techniques, especially those using Deep Neural Networks (DNNs). Previous studies (e.g., [1], [2]) trained a DNN by using a large amount of labeled data to acquire a distance metric useful for matching, comparing, or ranking of the data. Feature representations extracted from the supervisedly trained DNNs yield retrieval accuracy significantly higher than traditional handcrafted features. However, collecting sufficient number of labeled data is often impractical since annotation by humans is quite laborious. Insufficiency in number of labeled data hampers the supervised learning of feature representations and their distance metric.

Recently, unsupervised learning of feature representations, which tries to obtain expressive features from a collection of unlabeled data, has received much attention. In general, unsupervised learning is more challenging than supervised learning since direct supervision by labels is not available. To train feature extraction, or encoder, DNNs without using labels, "pretext" tasks are usually employed. In the field of 2D image analysis, such pretext tasks as self-reconstruction [3], context prediction [4], pseudo label classification [5], and feature contrast [6] have been proposed. Training using these pretext tasks achieved a certain level of success in learning feature representations in an unsupervised manner.

We argue that these previous pretext tasks for unsupervised representation learning have two shortcomings. (i) First, features learned by most of the previous pretext tasks for classification [4]–[16] are not optimal for retrieval.

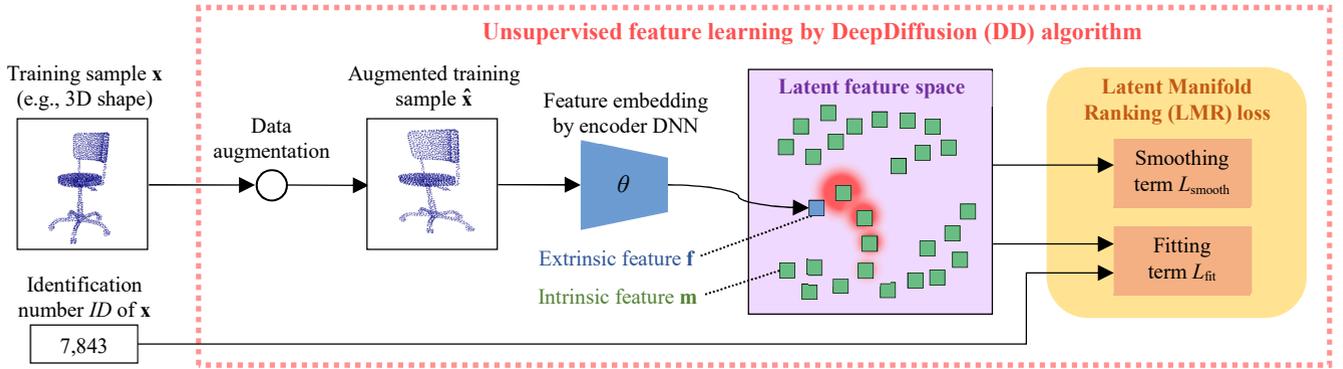

**FIGURE 1.** The proposed DeepDiffusion (DD) algorithm learns retrieval-adapted feature representations via ranking on a latent feature manifold. By minimizing Latent Manifold Ranking loss, the encoder DNN *and* the latent feature manifold (i.e., intrinsic features) are optimized for comparison of data samples. DD is applicable to a wide range of multimedia data including 3D shape and 2D image.

That is, DNN training using these pretext tasks attempts to form the latent feature space where similar data samples are embedded close to each other. However, none of the existing methods [4]–[16] impose any constraints on the structure of feature distribution in the learned latent feature space. Features learned without such constraints are expected to be nonlinearly distributed, possibly with gaps, in the latent feature space. Accurate classification can be achieved if decision boundaries among such nonlinearly distributed set of features are properly established by a subsequent supervised nonlinear classifier. On the other hand, however, retrieval accuracy suffers if the latent feature space is highly nonlinear, for example with bunching and/or gaps in the latent feature distribution. Feature comparison for retrieval is typically done by using a fixed distance metric (e.g., the Euclidean distance). The fixed distance metric has difficulty in finding all the data samples relevant to a query in the latent feature space having nonlinear distribution. To achieve accurate retrieval, a latent feature space having a continuous and smooth distance metric adapted to the nonlinear distribution of the input data is essential.

(ii) Second, some of the previous pretext tasks (e.g., [3], [4]) lack versatility, as they can only be applied to a specific data type. For example, the context prediction task [4] works only for 2D images since it heavily depends on the structure of 2D image. Or, the self-reconstruction [3] requires a loss function in addition to a pair of an encoder DNN and a decoder DNN, all of which must be exclusively designed for each data type.

Our goal in this paper is to develop a datatype agnostic unsupervised learning framework that learns *retrieval-adapted* feature representation. To this end, we introduce *diffusion distance on latent feature manifold*.

We first motivate the use of diffusion distance on latent feature manifold. Manifold hypothesis [61] assumes that high-dimensional data samples or their feature descriptors tend to lie in the vicinity of a nonlinear, low-dimensional manifold. Before the DNN era, manifold learning techniques [21] have empirically shown their effectiveness in tasks such as dimensionality reduction, clustering, and retrieval of diverse data types. We believe that the manifold hypothesis is also applicable to latent feature spaces formed by randomly initialized DNNs designed for encoding various data types such as 2D image and 3D shape. We thus hypothesize that incorporating the idea of manifold learning into the unsupervised deep learning could achieve our goal, i.e., learning retrieval-adapted features of diverse data types. Following our hypothesis, we attempt to optimize an encoder DNN and its latent feature space by using a pretext task that respects the structure of nonlinear manifold lying in the latent feature space. Our pretext task leverages diffusion distance on feature manifold [45] as a distance metric. Diffusion distance on a feature manifold is a powerful metric for information retrieval [45]. Given a set of discrete data points, or nodes, diffusion distance is defined as an average length of multiple paths between two nodes on the feature manifold graph. Averaging distances through multiple paths produces a smoother distance metric among a set of nodes. As described in [45], diffusion distances are typically computed by iteratively propagating similarities on the manifold graph. Many studies on multi-media information retrieval (e.g., [17], [18], [19], [45]) demonstrated that the data-adaptive diffusion distance can retrieve semantically similar data more accurately than a non-data-adaptive, or fixed, distance such as Euclidean distance. These successes motivate us to use the diffusion distance on a feature manifold, which tends to be smooth, for learning retrieval-adapted features. Also, to achieve applicability to a wide range of multimedia data, our pretext task is formulated as a loss function so that it works in conjunction with various data types, datasets, and encoder DNN architectures.

Our motivation described above is realized as a novel algorithm called *DeepDiffusion* (DD). Fig. 1 illustrates the learning framework of the DD algorithm. The crux of the DD algorithm is its loss function, which is designed to learn retrieval-adapted features. Our loss function, named *Latent Manifold Ranking* (*LMR*) loss, is built by reformulating the loss function of the Manifold Ranking algorithm [17], [19].

As shown in Fig. 1, the LMR loss is computed based on diffusion-based ranking in the latent feature space. The LMR loss is used to *jointly optimize* two sets of parameters, that are, connection weights within the encoder DNN and the manifold formed by latent features of all training samples. This optimization aims to transform the initial latent feature space so that a submanifold consisting of mutually similar features to contract while mutually distinct submanifolds to move farther away from each other. Diffusion distance-based ranking is achieved by comparing the latent features using a fixed (e.g., Euclidean) distance in the transformed latent feature space. After a training, feature vectors extracted by the trained encoder DNN are used for retrieval. In addition to this simple feature extraction, we also propose another feature extraction method that exploits both the trained encoder and the optimized latent feature manifold to boost retrieval accuracy.

Strengths of the DD algorithm are threefold. First, it learns features appropriate for content-based matching, comparison, and ranking of multimedia data. Second, our method learns such retrieval-adapted features in a fully unsupervised manner. The DD algorithm requires no semantic labels for its training. Third, our algorithm is widely applicable to any type of multimedia data as long as encoder DNNs that process the multimedia data type are available. This is because the DD relies only on its loss function.

We evaluate the effectiveness of the DD algorithm by using two types of multimedia data, that are, 3D shape represented by point set and 2D image represented by pixels. The experiments demonstrate that feature representations learned by the DD algorithm yield retrieval accuracy significantly higher than those learned by the existing methods for unsupervised feature learning.

Contributions of this paper can be summarized as follows.

- Proposing a novel unsupervised representation learning algorithm named DeepDiffusion (DD). Training framework of the DD algorithm is suitable to learn retrieval-adapted features and is applicable to diverse multimedia data types.
- Empirically evaluating accuracy and versatility of the DD algorithm by using scenarios of 3D shape retrieval and 2D image retrieval.

The rest of the paper is organized as follows. Related work is reviewed in Section II and our proposed algorithm is described in Section III. Section IV reports experimental results. Finally, conclusion and future work are discussed in Section V.

## II. RELATED WORK

### A. DNN-BASED DISTANCE METRIC LEARNING

Supervised *deep metric learning* is a well-studied approach to improving accuracy of information retrieval [48]. Majority of the approach creates supervisory signals by forming pairs [1], triplets [2], or higher order tuples [54], of labeled training samples. A latent feature space is optimized so that intra-class distances become smaller than inter-class distances. Some studies [55], [56] exploit user-generated tags as weak supervisory signal for deep metric learning. Meanwhile, a group of studies [49], [50], [57] try to directly optimize retrieval accuracy indices such as Average Precision and Recall via DNN training. They adopt ranked lists as supervisory signals and train the DNN with a dedicated objective function to maximize the retrieval accuracy. Although these approaches ([48]–[50], [57]) are effective in improving retrieval accuracy, they require a large number of labeled data for training.

Unsupervised fine-tuning of a DNN is one of the alternatives for deep metric learning. [51], [52], [58], [59] mine pseudo supervisory signals from latent features of unlabeled data to fine-tune the DNN. Although the mining procedures are unsupervised, they heavily depend on latent features extracted by the DNN pre-trained with labeled data. In contrast, our approach is purely unsupervised. That is, a DNN is trained from scratch by using only unlabeled data to obtain latent features suitable for retrieval.

### B. DNN-BASED UNSUPERVISED REPRESENTATION LEARNING

Unsupervised learning can potentially leverage a large amount of unlabeled data as training samples. To effectively train DNNs without relying on labels, pretext tasks are usually employed. Various pretext tasks for unsupervised representation learning have been proposed mainly in the field of 2D image analysis.

Autoencoder (AE) [3] learns features via a self-reconstruction task. An AE is formed by paring an encoder DNN with a decoder DNN. The encoder embeds an input data into the latent feature space while the decoder reconstructs the input from the embedded feature. Generative Adversarial Network (GAN) [20] learns a sample distribution of training dataset through adversarial training of a generator DNN and a discriminator DNN. Although GAN was originally proposed to generate realistic 2D images, it has also proven useful for unsupervised visual feature learning [7]. The other pretext tasks for unsupervised visual feature learning include, for example, predicting different channels (e.g., [8]), predicting spatial context (e.g., [4]), and predicting geometric transformation (e.g., [9]) of 2D images. Though these pretext tasks perform reasonably well in learning feature representation, they lack versatility. Designing the pretext task requires deep understanding of the data type. In addition, designing a DNN architecture and/or a loss function for the data type also requires knowledge of the data type.

Compared to the abovementioned pretext tasks, pseudo-label classification (e.g., [5]) and feature contrast (e.g., [6]) are more versatile and flexible. These tasks do not require the decoder DNN, and the encoder DNN is trained with a loss function that can be used independently of data type. The pseudo-label classification approach automatically assigns a pseudo category label to each unlabeled training sample. Instance Discrimination [5], [10], [60] assigns a unique pseudo-label per training sample. The pseudo-labels of DeepCluster [11] are generated by using $k$-means clustering of a set of latent features extracted by an encoder DNN. Swapping Assignments between multiple Views (SwAV) [12] and Local Aggregation [13] incorporate a clustering procedure as a part of a DNN to generate pseudo-labels adapted to the training data. On the other hand, feature contrast, also known as contrastive learning, trains the encoder DNN by comparing latent features. The feature contrast approach tries to form a latent feature space where features of positive sample pairs are embedded closer while features of negative sample pairs are embedded further apart from each other. Majority of the feature contrast methods, e.g., [6], [14] and [15] creates a positive pair by coupling two different "views" of a training sample. These two views are generated by applying data augmentation with different augmentation parameters to the training sample. A negative pair is formed between the pair of views derived from mutually different training samples. Features learned by pseudo-label classification or feature contrast were experimentally proven to be effective for classification and segmentation of multimedia data, especially 2D images.

Nevertheless, the existing pretext tasks mentioned in this section do not necessarily learn features that are suitable for information retrieval. Training by these pretext tasks do not pay much attention to the structure of the initial latent feature manifold formed by a randomly initialized encoder DNN. In addition, these pretext tasks do not impose any constraints on the learned latent feature distribution. Therefore, the training would strongly distort the initial latent feature manifold and converges to an unconstrained latent feature space having nonlinear distribution. Such nonlinear latent feature distribution, possibly with folds, bunches and gaps, is unsuitable for information retrieval ranked by using a fixed distance metric. For example, pseudo-labels of DeepCluster [11] ignore the nonlinearity of latent feature manifold since they are created by the $k$-means clustering. If one of the estimated centroids lies among different submanifolds in the latent feature space, DeepCluster overly deforms these submanifolds so that their latent features agglomerate as one cluster. Also, positive/negative pairs used by the feature contrast approach [6], [14], [15] may distort the latent feature manifold since negative samples are chosen randomly from a training data set. If both positive and negative samples lie on the same submanifold, it would be torn apart to converge to a latent feature space that is not optimal for retrieval. Our DD, on the other hand, learns retrieval-adapted features based on the criterion entirely different from the existing methods, that is, diffusion distance on the latent feature manifold.

### C. GRAPH-BASED MANIFOLD LEARNING

Manifold learning is a powerful technique to discover a nonlinear low-dimensional subspace inherent in a high-dimensional feature space. Various manifold learning algorithms were proposed primarily before the DNN era [21]. To unveil an intrinsic geometry of the high-dimensional space, manifold learning algorithms analyze a manifold graph. In this graph, each node represents a sample (i.e., a raw datum or its feature vector) and a set of such nodes are connected by edges whose weights are calculated based on proximity of the samples. The representative manifold learning algorithms (e.g., [22], [23]) use eigenanalysis of the manifold graph to nonlinearly project the samples to the low-dimensional subspace. Manifold Ranking (MR) [17][19] learns a distance metric among the samples by using similarity diffusion on the manifold graph. We later review the MR algorithm.

These classical manifold learning algorithms described above have been widely used for compression, clustering, comparison, and visualization, of high-dimensional data. However, accuracy of the distance metric learned by these classical algorithms highly depends on quality of data samples associated with nodes. Analyzing the manifold graph formed by raw data or their handcrafted features often produces a sub-optimal distance metric.

Some studies attempt to combine the idea of manifold learning with deep learning [24]–[27] for "manifold-aware" DNNs. Each of these studies proposes either a loss function or a DNN architecture that considers the manifold structure of input features. [24]–[26] propose loss functions that learn a latent feature space whose distance metric is constrained by proximity, in an input feature space, of training samples. [27] proposes to use a graph convolution layer as a building block of a DNN to diffuse similarity over a manifold formed by input features.

These manifold-aware deep learning methods [24]–[27] succeeded in finding feature embedding better than the classical manifold learning methods [21]. However, most of the existing manifold-aware DNNs do not optimize feature extraction since they process handcrafted features extracted prior to training the embedding DNN. In contrast, we take a holistic approach: our proposed DD optimizes both extraction of features from raw input data *and* embedding of the extracted features to the latent feature space.

#### 1) Manifold Ranking algorithm

Here, we briefly review the Manifold Ranking algorithm [17], [19], whose loss function serves as the basis of our LMR loss. MR learns a distance metric for retrieval of diverse multimedia data (e.g., [18]). MR works either in supervised, semi-supervised, or unsupervised mode. We describe the unsupervised MR below.

Given a set of $N$ unlabeled feature vectors, MR first constructs a manifold graph $\mathbf{W}$ by connecting neighboring samples in the feature space. $\mathbf{W} \in \mathbb{R}^{N \times N}$ is an adjacency matrix whose element $w_{ij}$ represents a similarity between two samples $i$ and $j$. Usually, $w_{ij}$ is calculated from the Euclidean distance between the input feature vectors. The objective of MR is to find a set of ranking score vectors $\{\mathbf{r}_i\}_{i=1, ..., N}$ according to the diffusion distances on the manifold. The ranking vector $\mathbf{r}_i$ contains similarity values between query $i$ that serves as the diffusion source and all the other features lying on the manifold. The loss function associated with $\{\mathbf{r}_i\}_{i=1, ..., N}$ is formulated in (1). For simplicity, we omit some coefficients contained in the original loss function [19].

$$\underset{\{\mathbf{r}_1, ..., \mathbf{r}_N\}}{\operatorname{argmin}} \sum_{i=1}^{N} \|\mathbf{r}_i - \mathbf{y}_i\|^2 + \lambda \sum_{i=1}^{N} \sum_{j=1}^{N} w_{ij} \|\mathbf{r}_i - \mathbf{r}_j\|^2 \qquad (1)$$

In (1), the first term is called fitting constraint, which fits the ranking score vector $\mathbf{r}_i$ of the query $i$ to the one-hot vector $\mathbf{y}_i$ representing the diffusion source. The second term is called smoothing constraint, which makes the query and its neighbors in the input feature space to have similar ranking scores. Eq. 1 can be solved either by an iterative form or a closed form solutions [17] [19]. Both solutions, however, require high spatial and temporal costs to analyze $N \times N$ adjacency matrix when the number of samples $N$ is large (e.g., $N > 10$k). Our proposed loss function is more efficient and is compatible with DNN training on a larger dataset (e.g., $N \sim 100$k).

## III. PROPOSED METHOD

### A. OVERVIEW OF DEEPDIFFUSION ALGORITHM

Fig. 1 illustrates our DeepDiffusion (DD) algorithm. The concept of the DD algorithm is to incorporate the idea of the MR algorithm, which leverages smooth diffusion distance on feature manifold for retrieval of diverse data types, into the problem of unsupervised deep representation learning. DD differs significantly from the existing pretext tasks for unsupervised deep representation learning in that DD respects the structure of a latent feature manifold formed by a randomly initialized encoder DNN. During a training by DD, both feature extraction and embedding by the encoder DNN are optimized so that the features are salient and the metric space is continuous and smooth. In the optimized latent feature space, the fixed distance (e.g., Euclidean distance) among the latent features approximates the diffusion distance on the latent feature manifold formed by the initial latent features.

Learning such retrieval-adapted latent feature space is achieved primarily by a novel loss function called Latent Manifold Ranking (LMR) loss. The LMR loss is built upon the loss function of the MR algorithm shown in (1). As depicted in Fig. 1, the LMR loss is computed by using two kinds of features, namely, *extrinsic* features and *intrinsic* features. An extrinsic feature is a feature of the input sample embedded by using the encoder DNN. An intrinsic feature is a feature lying on the manifold formed by latent features of all the training samples. The intrinsic features are not computed by the encoder, but updated via training.

One can use the MR loss shown in (1) as-is as a loss function for DNN training. However, directly applying the MR loss to DNN training confronts two issues. First, constructing a full manifold graph having $N^2$ connections from $N$ training samples suffers from high temporal and spatial computation costs. Our LMR loss mitigates this issue by constructing an ad-hoc, small manifold graph constructed from a subset of data samples for each training iteration. That is, at every training iteration, a bipartite manifold graph is formed between $B$ ($B \ll N$) extrinsic features extracted from an input minibatch and $N$ intrinsic features of all training samples. The use of the small manifold graph significantly reduces both temporal and spatial complexities compared to the full manifold graph. Second, the (squared) Euclidean distance used in the MR loss is not necessarily the best criterion for deep learning of distance metric. Recent studies on supervised deep metric learning (e.g., [28]) have demonstrated that comparing latent features by using divergence leads to a latent feature space better than that learned by using the Euclidean distance (e.g., [1], [2], [54]). Following the success of supervised deep metric learning, we employ divergence between probability distributions, instead of the Euclidean distance used in (1), to compare the ranking score vectors. Detail of the LMR loss is described in Section III.B.

After training, the optimized encoder DNN and the optimized latent feature manifold (i.e., intrinsic features) are used to extract features from novel data unseen during the training. We propose two feature extraction methods. The first method simply uses the trained encoder only. The second method exploits both the encoder and the latent feature manifold for feature extraction. Details of the feature extraction are described in Section III.C.

Since the crux of the DD algorithm is the LMR loss function, DD can potentially work in conjunction with any data type and encoder DNN architecture. As mentioned in Section I, we assume that the manifold hypothesis is valid in the latent feature space of an encoder DNN. If the initial latent features lie on a nonlinear manifold, DD is expected to learn retrieval-adapted latent features regardless of input data type. In the experiments, we demonstrate the versatility of DD by using multiple data types, datasets, and encoder DNN architectures.

### B. LATENT MANIFOLD RANKING LOSS

This subsection elaborates on the LMR loss, which is the core of the DD algorithm. We first define the symbols necessary to formulate our loss. Let $\{(\mathbf{x}_n, ID_n)\}_{n=1, ..., N}$ be a training dataset containing $N$ unlabeled samples $\mathbf{x}$. Each training sample **x** is paired with its unique identification number *ID*, which is used to specify a diffusion source. *ID* for the *n*-th training sample is *n*. In other words, the value of *ID* differs per training sample and does not change throughout training. The DD algorithm at the training stage takes as its input a mini-batch containing *B* training samples, i.e., $\{(\mathbf{x}_b, ID_b)\}_{b=1, ..., B}$. To diversify the training samples, we apply data augmentation to each sample **x** in the mini-batch to obtain $\hat{\mathbf{x}}$. The augmented sample $\hat{\mathbf{x}}$ is then embedded to the *P*-dimensional (e.g., *P*=256) latent feature space by using the encoder DNN parameterized by $\theta$. The embedded feature, or extrinsic feature, is denoted by **f**, whose *L2* norm is normalized to 1. Let $\mathbf{M} \in \mathbb{R}^{N \times P}$ be a matrix representing the latent feature manifold formed by all the training samples. **M** is constructed by stacking a set of *N* row vectors, i.e., $\{\mathbf{m}_n\}_{n=1, ..., N}$. $\theta$ and **M** are optimization targets of the DD algorithm.

Our objective function using the LMR loss is defined as (2), (3), and (4). Similar to the original MR loss (1), our LMR loss (2) comprises the fitting term $L_{\text{fit}}$ and the smoothing term $L_{\text{smooth}}$. The coefficient $\lambda$ balances these two terms.

$$\underset{\theta, \mathbf{M}}{\operatorname{argmin}} L_{\text{fit}} + \lambda L_{\text{smooth}} \quad (2)$$

$$L_{\text{fit}} = \sum_{b=1}^{B} H(\mathbf{r}_b, \mathbf{y}_b) \quad (3)$$

$$L_{\text{smooth}} = \sum_{b=1}^{B} \sum_{n=1}^{N} w_{bn} D(\mathbf{r}_b \| \mathbf{r}_n) \quad (4)$$

**Fitting term:** $L_{\text{fit}}$ constrains the ranking vector $\mathbf{r}_b$ of the training sample in the mini-batch to be close to the diffusion source vector $\mathbf{y}_b$. $\mathbf{r}_b$ is computed as $\mathbf{r}_b = \text{softmax}(\mathbf{f}_b \cdot \mathbf{M}^T)$. The *N*-dimensional vector $\mathbf{r}_b$ holds probabilistic similarities among the feature $\mathbf{f}_b$ and all the intrinsic features contained in **M**. $\mathbf{y}_b$ is an *N*-dimensional one-hot vector whose $ID_b$-th element is 1 while the other elements are 0. We use cross-entropy $H(\mathbf{r}_b, \mathbf{y}_b)$ to compare the ranking score vector and the diffusion source vector.

The effect of the fitting term is as follows. By minimizing $L_{\text{fit}}$, all the extrinsic features are embedded farther from each other since their ranking vectors are pulled toward different diffusion source vectors. Therefore, the fitting term is effective in learning salient features that can distinguish each training sample from other samples contained in the training dataset.

**Smoothing term:** $L_{\text{smooth}}$ constrains the extrinsic features and their neighboring intrinsic features to have similar ranking score vectors. Identical to the fitting term, $\mathbf{r}_b$ of $L_{\text{smooth}}$ is computed as $\mathbf{r}_b = \text{softmax}(\mathbf{f}_b \cdot \mathbf{M}^T)$. $\mathbf{r}_n$ is defined as $\mathbf{r}_n = \text{softmax}(\mathbf{m}_n \cdot \mathbf{M}^T)$ where $\mathbf{m}_n$ is *n*-th row of **M**. Thus, $\mathbf{r}_n$ contains ranking scores from the intrinsic feature $\mathbf{m}_n$ to all the intrinsic features including $\mathbf{m}_n$ itself. $w_{bn}$ indicates similarity between the extrinsic feature $\mathbf{f}_b$ and the intrinsic feature $\mathbf{m}_n$. $w_{bn}$ can be interpreted as one of the connection weights of the bipartite manifold graph, whose matrix size is *B*×*N*, formed between the *B* extrinsic features and the *N* intrinsic features. The value for $w_{bn}$ is computed by using (5), where $kNN(\mathbf{f}_b)$ is a set of *k* nearest intrinsic features of $\mathbf{f}_b$. We use Cosine distance in the latent feature space to compute $kNN(\mathbf{f}_b)$. The neighborhood size *k* is a hyper-parameter to be determined later. $D(\mathbf{r}_b \| \mathbf{r}_n)$ is a dissimilarity between the two ranking score vectors. We use the Jensen-Shannon divergence, i.e., a symmetric version of the Kullback-Leibler divergence, to measure the dissimilarity between the two probability distributions $\mathbf{r}_b$ and $\mathbf{r}_n$.

$$w_{bn} = \begin{cases} \mathbf{f}_b \cdot \mathbf{m}_n^T & \text{if } \mathbf{m}_n \in kNN(\mathbf{f}_b) \\ 0 & \text{otherwise} \end{cases} \quad (5)$$

By reducing $L_{\text{smooth}}$, the extrinsic features and their neighboring intrinsic features are attracted to each other in the latent feature space. In other words, when $L_{\text{smooth}}$ is small, the extrinsic features are likely to be projected onto the surface of the latent feature manifold formed by the intrinsic features. Therefore, concurrently minimizing $L_{\text{fit}}$ and $L_{\text{smooth}}$ helps the encoder DNN embed each training sample in the proximity of its corresponding intrinsic feature point on the latent manifold. In such a latent feature space, the distances between the extrinsic features computed as diffusion distances approximate the distances along the surface of the latent feature manifold.

**Choosing hyper-parameters:** The LMR loss has two hyper-parameters, that are, the balancing parameter $\lambda$ and the number of nearest intrinsic features *k*. Both hyper-parameters control smoothness of the feature distribution in the latent feature space. Optimal values for $\lambda$ and *k* would depend on various conditions including data samples used for training and the architecture of an encoder DNN. To validate generalization ability of the DD algorithm, we fix $\lambda$ at 1 and *k* at 20 when we empirically compare DD against the existing unsupervised feature learning algorithms. We also experimentally investigate the influences that these hyper-parameters have on retrieval accuracy.

**Initializing learnable parameters:** We randomly initialize the set of connection weights $\theta$ of the encoder DNN by using the algorithm proposed by He et al. [29]. Subsequently, each of the training samples is projected into a feature in the latent space by using the randomly initialized encoder. These randomly projected feature vectors are used as the initial values of the latent feature manifold matrix **M**. We employ such initialization of **M** since several studies [30][31] have shown that randomly initialized DNNs extract features having some degree of accuracy. We expect that **M** initialized with the randomly projected features represents more accurate initial manifold structure than **M** initialized with random values sampled from, for example, a normal distribution. The effectiveness of initializing **M** with randomly projected features will be shown in the experiments.

**Effect of data augmentation:** Data augmentation is applied to each training sample prior to passing it to the DNN. We expect that the training with data augmentation yields latent features robust against perturbation of training samples (e.g., affine transformation of 3D shapes). The fitting term $L_{fit}$ facilitates learning such robust features. That is, ranking vectors of multiple augmented data samples derived from an original training sample are pulled toward the same diffusion source vector as their original. As a result, latent features become less sensitive to the perturbation caused by the data augmentation. The detailed procedure of the data augmentation is described in Section IV.A.

Algorithm 1 summarizes the procedures of unsupervised DNN training by the DD algorithm.

### C. FEATURE EXTRACTION

After the training, the encoder DNN and the latent manifold are used for feature extraction from unseen data samples. We propose two features, that are, an embedded feature DD (E) and a diffused feature DD (D). Computation of the embedded feature is straightforward. An input sample is projected to the latent feature space by using the encoder DNN. The output from the encoder, i.e., **f**, is the DD (E) feature of the input. The Euclidean distance between the pair of DD (E) features is expected to approximate the diffusion distance on the latent feature manifold. Algorithm 2 summarizes the procedures for extracting the DD (E) feature.

We observe that the iterative diffusion process adopted by the traditional diffusion-based ranking algorithms [45] would make the DD (E) feature better fit for retrieval. We thus propose the DD (D) feature, which is a smoothed version of the DD (E) feature. The DD (D) feature is computed by similarity diffusion on the latent feature manifold. After computing **f**, we find its $k$ nearest intrinsic features $kNN(\mathbf{f})$ from the optimized **M**, or $\{\mathbf{m}_n\}_{n=1,...,N}$, to determine multiple diffusion sources. $kNN(\mathbf{f})$ is a set of $k$ intrinsic features closest to **f** found by using the Euclidean distance. The diffusion sources are represented by an $N$-dimensional vector $\mathbf{g}_0$. The $n$-th element of $\mathbf{g}_0$, i.e., $g_0(n)$, is computed by using (6).

$$\mathbf{g}_0(n) = \begin{cases} 1 & \text{if } \mathbf{m}_n \in kNN(\mathbf{f}) \\ 0 & \text{otherwise} \end{cases} \quad (6)$$

We use the recursive formula $\mathbf{g}_r = \mathbf{g}_{r-1} \cdot \mathbf{S}$, which is one of

---

**Algorithm 1** Training by using the DD algorithm.

1: **Inputs:** The training dataset $\{(\mathbf{x}_n, ID_n)\}_{n=1,...,N}$
2:     The encoder DNN parameterized by $\theta$
3:     The balancing coefficient $\lambda$ (e.g., $\lambda=1$)
4:     The number of neighbors $k$ (e.g., $k=20$)
5:     The optimization algorithm (e.g., Adam [43] with initial learning rate $\eta=10^{-4}$)
6: **Outputs:** Optimized encoder parameters $\theta$
7:     Optimized intrinsic features **M**, or $\{\mathbf{m}_n\}_{n=1,...,N}$, of the training samples
8: **Initialization:**
9:     Randomly initialize $\theta$.
10:    Initialize intrinsic features **M** by feeding all the training samples into the encoder DNN.
11: **for** each training epoch **do**
12:    **for** each minibatch $\{(\mathbf{x}_b, ID_b)\}_{b=1,...,B}$ sampled from the training dataset **do**
13:       Augment the samples $\{\mathbf{x}_b\}_{b=1,...,B}$ to obtain $\{\hat{\mathbf{x}}_b\}_{b=1,...,B}$.
14:       Feed the augmented samples $\{\hat{\mathbf{x}}_b\}_{b=1,...,B}$ into the encoder DNN to extract their latent features $\{\mathbf{f}_b\}_{b=1,...,B}$.
15:       Create one-hot vectors $\{\mathbf{y}_b\}_{b=1,...,B}$ from $\{ID_b\}_{b=1,...,B}$.
16:       Compute $L_{fit}$ by using $\{\mathbf{f}_b\}$ and $\{\mathbf{y}_b\}$ as inputs to Eq. (3).
17:       Compute $L_{smooth}$ by using $\{\mathbf{f}_b\}$, **M**, and $k$ as inputs to Eq. (4) and (5).
18:       Compute the overall loss value: $L \leftarrow L_{fit} + \lambda L_{smooth}$.
19:       Update $\theta$ and **M** by using the optimization algorithm:
20:         $\theta \leftarrow \theta - \eta \frac{\partial L}{\partial \theta}$
21:         $\mathbf{M} \leftarrow \mathbf{M} - \eta \frac{\partial L}{\partial \mathbf{M}}$
22:    **end for**
23: **end for**
24: **return** $\theta$ and **M**

---

**Algorithm 2** Extracting DD (E) feature.

1: **Inputs:** The data sample **x**
2:     The encoder DNN optimized by Algorithm 1
3: **Output:** The DD (E) feature **f** of **x**
4: Feed the input sample **x** into the encoder DNN to obtain the latent feature **f**.
5: **return f**

---

**Algorithm 3** Extracting DD (D) feature.

1: **Inputs:** The data sample **x**
2:     The encoder DNN optimized by Algorithm 1
3:     The intrinsic features **M** optimized by Algorithm 1
4:     The number of neighbors $k$ (e.g., $k=20$)
5:     The number of iterations $R$ for similarity diffusion (e.g., $R=20$).
6: **Output:** The DD (D) feature $\mathbf{g}_R$ of **x**
7: Construct the sparse similarity matrix **S**: (**S** can be precomputed prior to feature extraction.)
8:     $\mathbf{S} \leftarrow \mathbf{M} \cdot \mathbf{M}^T$
9:     **for** each row $\mathbf{s}_i$ of **S do**
10:       Sparsify $\mathbf{s}_i$ by replacing all the elements except for the $k$ largest elements with 0.
11:    **end for**
12: Feed the input sample **x** into the encoder DNN to obtain the latent feature **f**.
13: Find a set of $k$ nearest intrinsic features of **f**, i.e., $kNN(\mathbf{f})$, from the intrinsic features **M**.
14: Create the diffusion source vector $\mathbf{g}_0$ by using Eq. (6).
15: **for** $r = 1, ..., R$ **do**
16:     Diffuse similarity: $\mathbf{g}_r \leftarrow \mathbf{g}_{r-1} \cdot \mathbf{S}$
17: **end for**
18: Normalize the scale of $\mathbf{g}_R$: $\mathbf{g}_R \leftarrow \mathbf{g}_R / \|\mathbf{g}_R\|$
19: **return** $\mathbf{g}_R$

the simplest forms of diffusion computation on a feature manifold [45], to derive similarity. $\mathbf{S} \in \mathbb{R}^{N \times N}$ is a sparse similarity graph whose $(i, j)$ element is $\mathbf{m}_i \cdot \mathbf{m}_j^T$ if $\mathbf{m}_j$ is included in a set of $k$ nearest neighbors of $\mathbf{m}_i$, and 0 otherwise. By iterating the recursive formula, the diffusion source vector $\mathbf{g}_r$ becomes smooth according to the pairwise similarities of the intrinsic features optimized during the training. We iterate the recursion for $R$=20 times, whose validity is evaluated in the experiments. After the iteration, $\mathbf{g}_R$ is scaled to unit length to obtain the DD (D) feature. $k$ for computing the diffusion feature is identical to $k$ used in the LMR loss. Algorithm 3 summarizes the procedures for extracting the DD (D) feature.

Retrieval rankings are generated by using the Euclidean distances among either the DD (E) features, the DD (D) features, or the "fused" features. Intending to improve retrieval accuracy, the fused feature DD (E+D) is formed by concatenating the DD (E) vector (i.e., $\mathbf{f}$) and the DD (D) vector (i.e., the scaled $\mathbf{g}_R$).

## IV. EXPERIMENTS AND RESULTS

### A. EXPERIMENTAL SETUP

We comprehensively evaluate the effectiveness of the proposed DD algorithm through the experiments using multiple data types (i.e., 3D shape and 2D image), multiple datasets, and multiple encoder architectures.

**Datasets:** We use three 3D shape datasets and three 2D image datasets. Table 1 summarizes the statistics for each dataset. ModelNet10 (MN10) [32], ModelNet40 (MN40) [32], and ShapeNetCore55 (SN55) [33] contain 3D polygonal models of rigid 3D objects such as furniture, vehicles, and household appliances. We convert each 3D shape to a set of 1,024 3D points by using the algorithm by Ohbuchi [34]. Each 3D point set is normalized so that its gravity center coincides the origin of the 3D space and its whole shape is enclosed by a unit sphere. Fashion MNIST (FMNIST) [35] includes 2D images of clothes, while STL10 [37] consists of 2D natural images. COIL100 [36] contains 2D images of 100 different objects placed on a turntable. For each object, a set of 72 images was taken at intervals of 5 degrees while rotating the turntable. We use the images taken between 0 to 175 degrees for training, and the rest for evaluation.

Retrieval accuracy of the testing data is measured in Mean Average Precision (MAP) [%], which is the de facto standard accuracy index for information retrieval systems.

**Competitors:** The DD algorithm is compared against 11 existing unsupervised feature learning algorithms. They are AE [3], GAN [7], Noise As Targets (NAT) [16], Parametric Instance Discrimination (PID) [5], Non-Parametric Instance Discrimination (NPID) [10], DeepCluster (DC) [11], SwAV [12], LocalAggregation (LA) [13], Invariant and Spreading Instance Feature (ISIF) [6], Simple framework for Contrastive Learning of Representations (SimCLR) [14], and Momentum Contrastive learning (MoCo) [15]. These competitors train their encoder DNNs by using different training objectives, or loss functions. AE uses a self-reconstruction loss. We use the Chamfer distance loss [62] for the 3D point set AE and the mean squared error loss for 2D image AE. GAN uses the adversarial loss [7] for training. NAT, PID, NPID, DC, SwAV, and LA employ pseudo-label classification loss, while ISIF, SimCLR, and MoCo employ feature contrast loss. Pseudo-label classification and feature contrast are training objectives widely used for unsupervised feature learning. We fix the number of clusters for DC, SwAV, and LA at roughly 10 times the number of object categories, as suggested in [11].

**Implementation details:** For a fair comparison, we use the same data augmentation, DNN architecture, and optimization method among all the algorithms used in the experiments.

We employ online data augmentation. Each training sample in the mini-batch is augmented with a probability of 0.8. A 3D point set is augmented by random affine transformation. Specifically, the point set is first randomly rotated about the x-, y-, and z-axes in this order. The rotation angle for each axis is randomly and independently chosen from the uniform distribution $U(−5°, 5°)$. After the rotation, the point set is anisotropically scaled. Directions of scaling are the x-, y-, and z-axes, and scaling factor for each axis is randomly sampled from $U(0.8, 1.2)$. The 3D shape is then sheared successively in the x-, y- and z-axes. Shearing factor for each axis is sampled from $U(−0.2, 0.2)$. Finally, the 3D shape is randomly translated. Displacement along each axis is chosen from $U(−0.2, 0.2)$. A 2D image is augmented by random cropping and horizontal flipping. The 2D image is isotropically scaled by the factor 1.2, and a rectangle having the original image size is cropped at random position. The cropped image is then horizontally flipped with the probability of 0.5.

We use four popular DNN architectures as encoders. To encode 3D point sets, we adopt PointNet [38] and Dynamic Graph CNN (DGCNN) [39]. To process 2D images, we employ Inception V1 (also known as GoogLeNet) [40] and ResNet18 [41]. The number of dimensions $P$ of the latent space is fixed at 256 throughout the experiments. In addition to the encoder DNN, AE and GAN require a decoder DNN. We employ the 3D point set decoder composed of five fully-connected layers [42] and the 2D image decoder consisting

**Table 1** 3D shape/2D image datasets used in our experiments.

| Dataset | Data format for each sample | Number of categories | Number of training data | Number of testing data |
|---|---|---|---|---|
| MN10 [32] | a set of 1,024 3D points | 10 | 3,991 | 908 |
| MN40 [32] | | 40 | 9,843 | 2,468 |
| SN55 [33] | | 55 | 35,764 | 10,265 |
| FMNIST [35] | 28×28 pix., 1 ch. | 10 | 60,000 | 10,000 |
| COIL100 [36] | 128×128 pix., 3 ch. | 100 | 3,600 | 3,600 |
| STL10 [37] | 96×96 pix., 3 ch. | 10 | 100,500 | 800 |

of five deconvolution layers [7].

The DNNs are trained by using mini-batch gradient descent. We use Adam [43] with the initial learning rate of $10^{-4}$. Each mini-batch contains 16 3D shapes or 64 2D images. We iterate the training for 300 epochs and report the highest MAP scores obtained during the training.

The DD algorithm is implemented in Python using TensorFlow library [44]. Most of the experiments were done on a PC having an *Intel Core i9-7900X* CPU and an *Nvidia GeForce RTX 2080 Ti* with 11 GBytes of GPU memory.

### B. EXPERIMENTAL RESULTS AND DISCUSSION

#### 1) Comparison with existing algorithms

Table 2 compares retrieval accuracies of the 12 unsupervised feature learning algorithms including the proposed DD. In the tables, "DD (E)" and "DD (D)" indicate the embedded feature and the diffused feature, respectively, learned by using the DD algorithm. "DD (E+D)" denotes concatenation of the two feature vectors above. The hyper-parameters of DD, i.e., $k$ and $\lambda$, are fixed at 20 and 1, respectively. As the reference, Table 2 includes "untrained" that shows accuracy of the features extracted by using the randomly initialized encoder DNN.

As shown in Table 2, our DD algorithm yields MAP scores higher than the existing methods for almost all the combinations of the encoder DNNs and the datasets. These results verify that optimization using the diffusion distance on the latent feature manifold is effective for learning feature representation adapted to multimedia information retrieval.

In addition, Table 2 compares efficacy of the two proposed features, i.e., DD (E) and DD (D). The winner among DD (E) and DD (D) depends on dataset and encoder used. But the fused feature DD (E+D) slightly but consistently outperforms the two. We suspect that distances among the embedded features of the testing data slightly differ from the diffusion distances on the latent manifold since the testing data are not used in training. Thus, the diffused feature, which is computed by using the latent feature manifold, would assist the embedded feature to find larger number of true neighbors in the latent feature space.

Fig. 2 visualizes the latent feature spaces learned by using the three unsupervised feature learning algorithms, i.e., AE, SwAV, and our DD. We use PointNet and Inception V1 as encoder DNNs and use DD (E) as the latent features of DD. The t-SNE algorithm [53] is used to visualize the latent features extracted from the testing samples of the MN10 and FMNIST datasets. As can be observed in Fig. 2, the latent features learned by the DD algorithms are well-separated. That is, each cluster of the latent features consists of one or two semantic categories in most cases and these clusters are reasonably far from each other. Such feature embeddings contribute to the high retrieval accuracy of the DD algorithm shown in Table 2. On the other hand, AE and SwAV tend to learn feature embeddings that are not suitable for information retrieval. In the latent space of AE, the feature clusters overlap each other. SwAV learns favorable feature embeddings for the MN10 dataset, but the latent features of the FMNIST dataset are separated into small clusters. Since the training frameworks of AE and SwAV are not intended to acquire retrieval-adapted features, they do not necessarily form latent feature spaces suitable for retrieval.

#### 2) In-depth evaluation of DD algorithm

This subsection investigates the influences that various design parameters of the DD algorithm have on retrieval accuracy. We use the MN10, MN40, FMNIST, and COIL100 datasets for evaluation. PointNet and Inception V1 are employed as encoder DNNs.

**Hyper-parameters:** Fig. 3 plots retrieval accuracies of the DD (E) feature against the number of nearest intrinsic features $k$ used for computing the smoothing term. We can observe that, for all the four datasets, MAP scores gradually improve with increasing $k$ from 1 to 20. Optimal $k$ depends on dataset (and probably also on encoder DNN). The FMNIST dataset requires larger $k$ than the other datasets to attain peak accuracy. This is probably because the category size of FMNIST (6,000 samples per category) is larger than those of the other datasets (tens to hundreds samples per category). When the category size is large, computing the smoothing term with large $k$ would facilitate capturing the manifold structure composed of many samples of the same category.

**Table 2** Comparison of retrieval accuracies (MAP [%]) using multiple data types, datasets, and encoder DNN architectures.

| Training objective | Unsupervised feature learning algorithm | 3D point set data | | | | | | 2D image data | | | | | |
|---|---|---|---|---|---|---|---|---|---|---|---|---|---|
| | | PointNet [38] encoder | | | DGCNN [39] encoder | | | Inception V1 [40] encoder | | | ResNet18 [41] encoder | | |
| | | MN10 dataset | MN40 dataset | SN55 dataset | MN10 dataset | MN40 dataset | SN55 dataset | FMNIST dataset | COIL100 dataset | STL10 dataset | FMNIST dataset | COIL100 dataset | STL10 dataset |
| − | untrained | 53.3 | 35.9 | 39.8 | 37.0 | 24.8 | 32.0 | 23.4 | 57.4 | 12.4 | 24.7 | 53.8 | 11.2 |
| self-reconstruction | AE [3] | 67.3 | 46.4 | 51.2 | 68.0 | 47.2 | 51.6 | 44.2 | 69.1 | 14.8 | 42.8 | 68.1 | 14.8 |
| data generation vs. discrimination | GAN [7] | 67.0 | 48.8 | 52.9 | 58.1 | 49.4 | 50.9 | 28.9 | 69.3 | 13.7 | 36.2 | 63.8 | 13.1 |
| Pseudo-label classification | NAT [16] | 68.5 | 47.8 | 50.1 | 59.0 | 47.2 | 48.9 | 37.9 | 80.7 | 16.1 | 46.8 | 83.0 | 16.8 |
| | PID [5] | 70.9 | 49.4 | 53.9 | 66.2 | 50.7 | 52.9 | 35.3 | 81.4 | 15.2 | 44.9 | 83.7 | 15.3 |
| | NPID [10] | 71.5 | 45.3 | 57.0 | 72.9 | 47.7 | 58.9 | 39.0 | 76.1 | 16.6 | 37.1 | 79.8 | 16.0 |
| | DC [11] | 70.4 | 46.1 | 54.9 | 71.6 | 48.1 | 53.6 | 43.5 | 84.3 | 18.6 | 39.5 | 86.3 | 16.2 |
| | SwAV [12] | 73.9 | 48.4 | 56.2 | 73.7 | 53.5 | 57.7 | 35.2 | 78.3 | 16.3 | 31.9 | 80.7 | 17.1 |
| | LA [13] | 76.2 | 54.0 | 44.4 | 74.9 | 57.2 | 43.9 | 31.9 | 80.7 | 17.1 | 34.3 | 81.2 | 16.5 |
| Latent feature contrast | ISIF [6] | 67.0 | 51.1 | 51.6 | 64.5 | 52.1 | 51.3 | 31.1 | 77.2 | 14.2 | 32.9 | 72.4 | 15.0 |
| | SimCLR [14] | 70.1 | 54.2 | 55.2 | 70.5 | 56.4 | 53.9 | 29.6 | 76.6 | 17.0 | 36.4 | 74.2 | 16.0 |
| | MoCo [15] | 70.3 | 52.8 | 52.2 | 67.8 | 54.3 | 51.9 | 35.2 | 81.4 | 16.7 | 37.3 | 82.5 | 17.2 |
| Similarity diffusion on latent feature manifold | DD (E) (ours) | **80.4** | **58.0** | 58.6 | **80.6** | **61.2** | 59.3 | 49.8 | 89.2 | **20.1** | 49.8 | 85.8 | **17.4** |
| | DD (D) (ours) | 76.8 | 56.2 | **61.4** | 76.5 | 58.2 | **61.7** | **50.0** | **89.3** | 19.7 | **50.9** | 84.1 | 17.3 |
| | DD (E+D) (ours) | **80.6** | **59.9** | **61.7** | **80.9** | **62.6** | **62.2** | **50.2** | **90.4** | **20.6** | **51.7** | **86.0** | **17.6** |

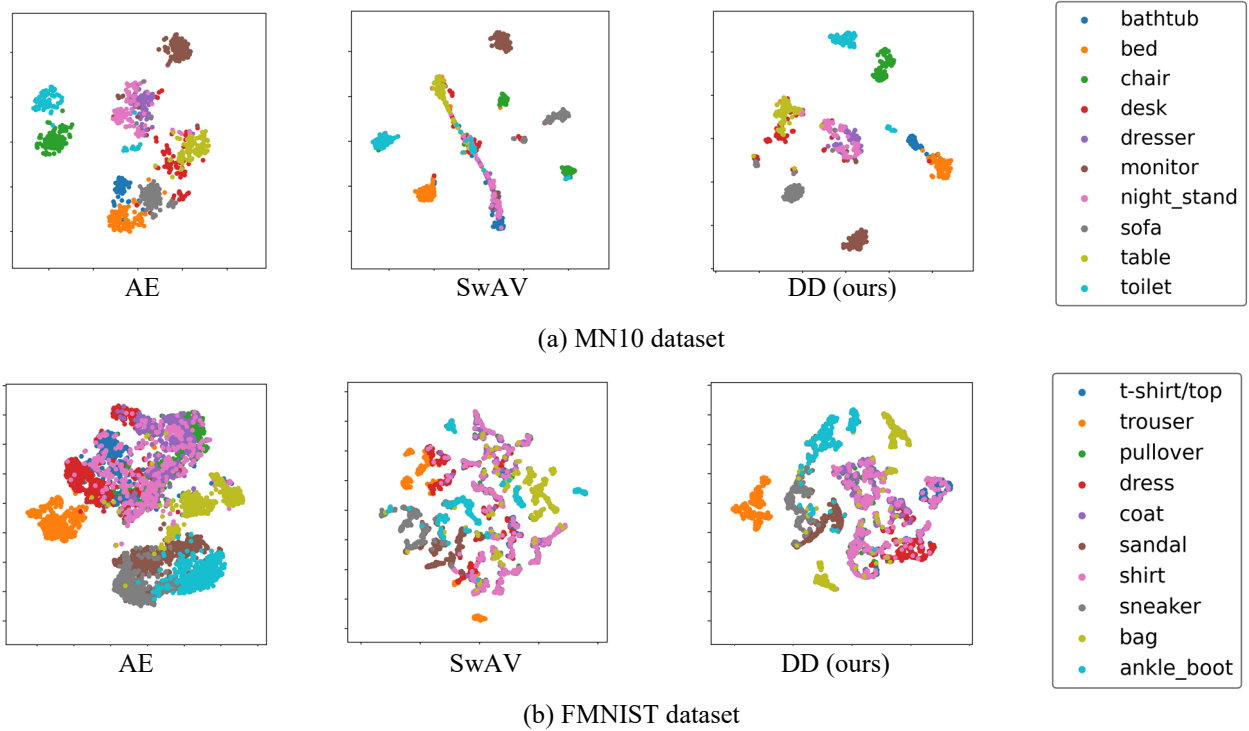

(a) MN10 dataset

(b) FMNIST dataset

**FIGURE 2** Visualization of the latent feature space learned by AE [3], SwAV [12], and our DD algorithm.

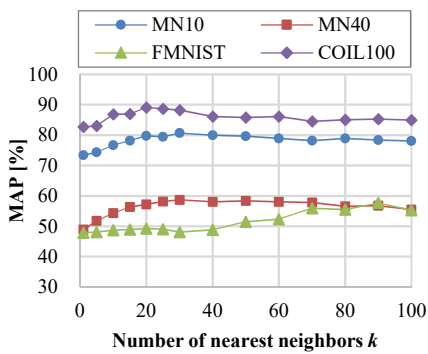

**FIGURE 3** Retrieval accuracy of the DD (E) feature plotted against the number $k$ of neighboring intrinsic features.

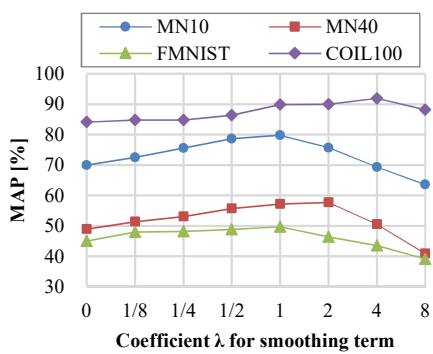

**FIGURE 4** Retrieval accuracy of the DD (E) feature plotted against the balancing parameter $\lambda$ for the LMR loss.

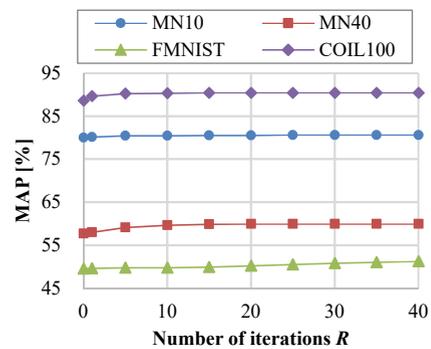

**FIGURE 5** Retrieval accuracy of the DD (E+D) feature plotted against the number of iterations $R$ for the diffused feature.

Fig. 4 shows the influence the balancing parameter $\lambda$ of the LMR loss has on the retrieval accuracy of the DD (E) feature. In the graph, $\lambda=0$ means the DNN is trained by using the fitting term only. Fig. 4 indicates that the smoothing term has positive impact on retrieval accuracy. However, excessively large $\lambda$ leads to low retrieval accuracy since gradients derived from the loss function is dominated by the smoothing term.

Fig. 5 shows retrieval accuracy of the DD (E+D) feature plotted against the number of iterations $R$ for calculating the diffused feature. Although the influence by $R$ on the accuracy is small, the diffusion with sufficient iterations slightly improves the MAP score. The accuracies saturate at about about $R=20$ for the MN10, MN40, and COIL100 datasets. In the FMNIST, the accuracy continues to improve at $R > 20$.

Since FMNIST has more training data than the other three datasets, more iterations would be required for convergence of the similarity diffusion.

**Ablation study:** We evaluate contributions from the components of the DD algorithm, that are, the loss function, the data augmentation of training samples, and the initialization of intrinsic features. Table 3 demonstrates that all the four components positively impact retrieval accuracy. The two terms of the LMR loss, i.e., $L_{fit}$ and $L_{smooth}$, are indispensable to learn retrieval-adapted features. The data augmentation clearly improves retrieval accuracy. Diversifying training samples would help the DD algorithm obtain robustness against perturbation of the samples and improve generalization ability.

Table 3 also shows that the initialization method for intrinsic features has a significant impact on accuracy. In the column "RPF init.", "Yes" indicates that the intrinsic features are initialized with the features of training data extracted by the encoder having random parameters. "No" means that the intrinsic features are initialized with randomly sampled values [29]. Evidently, initialization with random projection features improves accuracy. Fig. 6 compares training behavior between the two initialization methods. Initialization with random values appears to cause overfitting; MAP of the testing data saturates early (around 120 epochs) while loss value of the training data continues to decrease. In contrast, the training started with randomly projected features is more regularized and yields higher MAP score (nearly 80%) on the MN10 dataset.

**Computational cost for training:** We evaluate training scalability of the DD algorithm. To do so, we create pseudo large-scale training datasets by duplicating the 3D shape data contained in the MN10 dataset. During training, we measure GPU memory footprint and time per epoch. Fig. 7 shows that the training by DD is reasonably efficient when the number of training samples $N$ is less than 100K. However, both spatial and temporal costs grow significantly as $N$ increases. Using the GPU having 11 GBytes of memory, training is not executable at $N > 330K$ due to an out-of-GPU-memory error. Training by DD requires $O(N)$ spatial complexity since multiple $N$-dimensional ranking score vectors as well as $N$ intrinsic features need to be stored to compute the LMR loss. The increase in GPU memory usage against $N$, however, is not linear in Fig. 7. This is probably due to the behavior specific to the TensorFlow library. In terms of training time, the DD algorithm requires $O(N^2)$ temporal complexity since the similarities (i.e., $w_{bn}$ in (4)) of all pairs of the $N$ extrinsic features and the $N$ intrinsic features must be calculated during training of one epoch.

The high training costs of the DD algorithm might be mitigated by using subsampling methods (e.g., [46]) to reduce $N$ for loss computation, or approximate nearest neighbor search methods (e.g., [47]) to avoid the all-pair comparison among the extrinsic/intrinsic features. We leave for future work applications of these techniques to improve efficiency of the DD algorithm.

**Table 3** Ablation study of the DD algorithm. Accuracy of DD (E) feature is measured in MAP [%]. (DA: data augmentation, RPF init.: initializing intrinsic features with randomly projected features.)

| $L_{fit}$ | $L_{smooth}$ | DA | RPF init. | MN10 | MN40 | FMNIST | COIL100 |
|---|---|---|---|---|---|---|---|
| Yes | Yes | Yes | Yes | **80.4** | **58.0** | **49.8** | **89.2** |
| No | Yes | Yes | Yes | 56.4 | 35.7 | 20.9 | 63.7 |
| Yes | No | Yes | Yes | 70.0 | 48.9 | 45.0 | 84.1 |
| Yes | Yes | No | Yes | 71.2 | 49.2 | 47.2 | 81.7 |
| Yes | Yes | Yes | No | 75.8 | 53.3 | 32.5 | 87.8 |

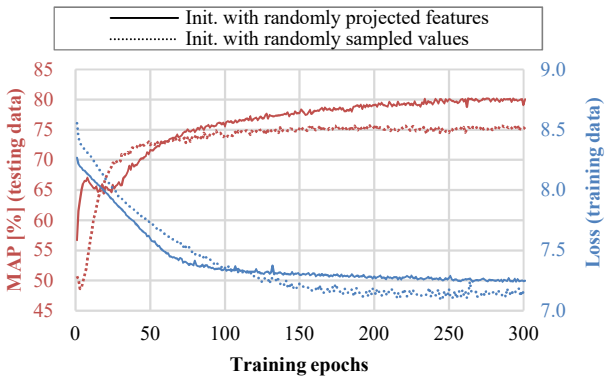

**FIGURE 6** Training behavior on the MN10 dataset using different initialization methods for the intrinsic features M.

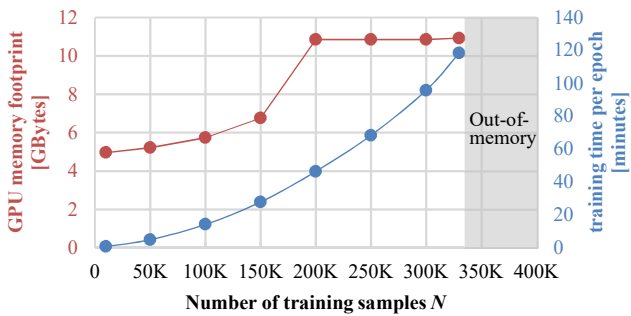

**FIGURE 7** Computation costs for DNN training as a function of the number of training samples $N$.

## V. DISCUSSION

While the quantitative experiments in Section IV show the effectiveness of the proposed DD algorithm, it is still unclear how DD behaves differently from the existing unsupervised deep feature learning algorithms. As we mentioned in Section I, DD aims to transform the initial latent feature space so that a submanifold consisting of mutually similar features to contract while mutually distinct submanifolds to move farther away from each other. To prove that DD works as intended, this section compares training behaviors of the multiple feature learning algorithms including DD. We visualize and observe changes in the manifold structure in the latent feature space during training. However, the visualization methods for high-dimensional space such as t-SNE used in Fig. 2 are not suitable for observing the true structure of feature manifold. This is because these visualization methods by themselves use some form of dimension reduction that could obscure the observation. We thus use a far simpler problem, i.e., a 2D toy dataset, to observe manifold structures. Our toy dataset consists of 1,000 2D points lying on the three-arm spiral distribution (see "0th epoch" in Fig. 8a).

We first describe the setup of the experiment. We use an encoder DNN consisting of four fully-connected layers. The input and output layers of the encoder have two neurons,

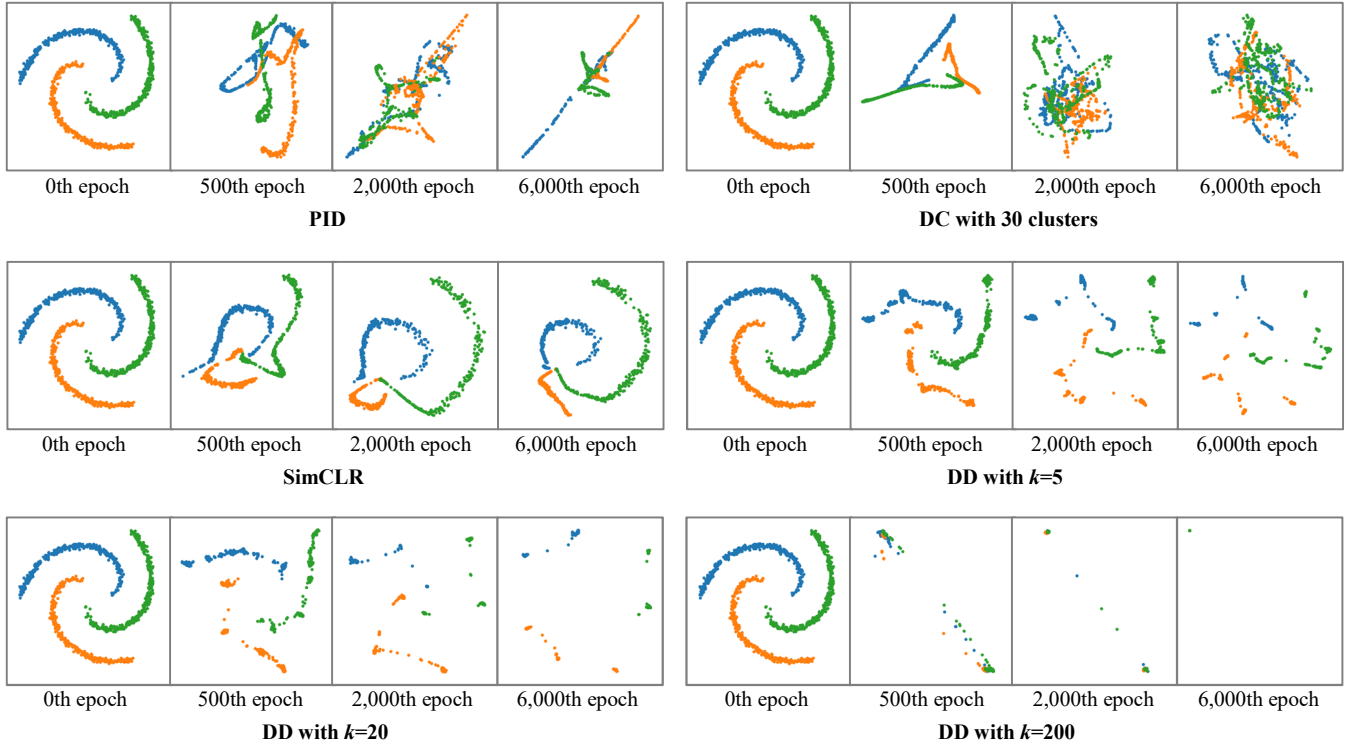

(a) Visualization of 2D latent feature space formed by 1,000 features extracted from the 2D toy data. Color of each latent feature point indicates its pseudo semantic category.

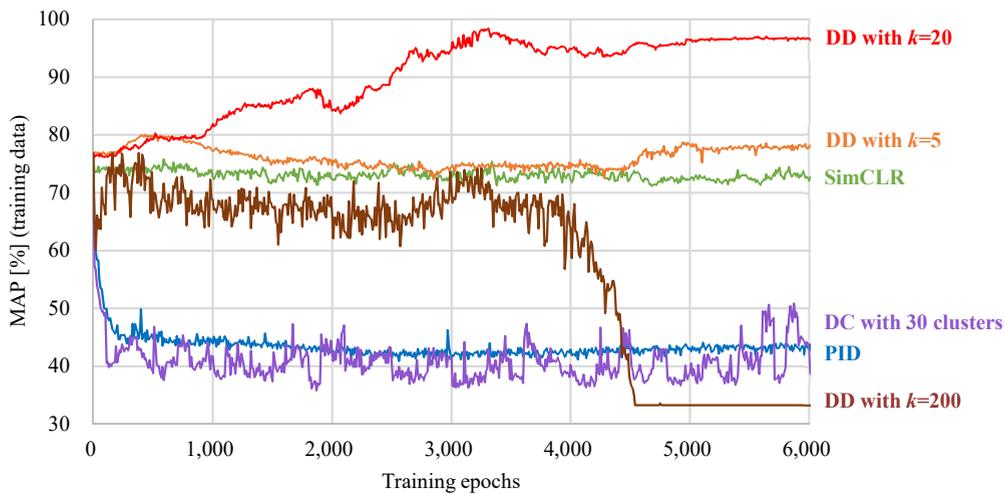

(b) Retrieval accuracy of 2D toy data plotted against training epoch.

**FIGURE 8.** Comparison of training behaviors of the unsupervised feature learning algorithms using the 2D toy dataset. (a) While the existing algorithms deform the latent feature space ignoring the structure of the initial latent feature manifold, our DD with the proper number of neighbors (*k*=20) succeeds in disentangling the nonlinearity of the initial latent manifold. (b) As the DNN training progresses, retrieval accuracy of DD with *k*=20 improves. That is, the latent features of DD with *k*=20 adapted to retrieval during training.

while the hidden layers have 1,000 neurons. Following the manifold hypothesis, we want to create a nonlinear feature manifold (e.g., the three-arm spiral) in the latent space of the encoder DNN. To do so, we pretrain the encoder DNN so that it learns an identity mapping. That is, we sample 10,000 2D points from the uniform distribution as training data for pretraining. The encoder DNN is pretrained so that the input 2D points are embedded in the same coordinates as the inputs in the 2D latent space. After the pretraining, the encoder DNN is further trained by using either PID, DC, SimCLR, or our DD. For DD, we use the different values for the number of neighbors $k$, i.e., $k$=5, 20, and 200. The 1,000 2D points lying on the three-arm spiral is used as a training dataset. The training is iterated for 6,000 epochs. In addition to observing the latent feature space, we compute retrieval accuracy of the latent features. To measure retrieval accuracy, we embed all the 1,000 2D toy data points in the training dataset and compare these latent features by using the Euclidean distance. A MAP score is computed assuming that the features lying on the same submanifold, i.e., the same arm of the spiral, belong to the same semantic category.

Fig. 8a visualizes changes in the 2D latent feature space during training. Fig. 8b shows retrieval accuracies, measured in MAP, during training. Note that the latent feature distribution and the MAP score (77%) at the beginning of training (0th epoch) are the same across all the methods in Fig. 8. Evidently, the six cases in Fig. 8 show different training behaviors. PID and DC destroy the structure of the initial latent feature manifold. That is, the three submanifolds during training overlap with each other, resulting in decrease in retrieval accuracy. PID in this experiment tries to embed the latent features in a straight line so that each training sample can be discriminated. However, PID fails to separate the green submanifold from the other two. DC in this experiment mixes up the latent features after 2,000 epochs. This is because pseudo labels, i.e., cluster centers, are generated among different submanifolds. Latent features are mixed up since those belonging to the different submanifolds are attracted to the same cluster center. For SimCLR, the latent feature space is hardly distorted and retrieval accuracy is almost unchanged throughout training. This is probably because the training objective of SimCLR, i.e., making distances among positive pairs smaller than distances among negative pairs, is achieved to some extent in the initial latent space. Although SimCLR does not show high retrieval accuracy, it is likely to yield high classification accuracy if a nonlinear classifier is trained in the learned latent feature space.

In contrast, DD behaves differently from PID, DC, and SimCLR. When DD uses an appropriate value for $k$ (i.e., DD with $k$=20), the nonlinearly distributed initial latent features are successfully disentangled. Although continuity of the initial feature distribution is lost to some degree, the three submanifolds at the 6,000th epoch are mutually separated and the latent features on the same arm are embedded close to each other. Retrieval accuracy of DD with $k$=20 improves as the training progresses, and MAP score reaches nearly 97%. These results indicate that the initial latent features properly adapt to retrieval via the training by DD.

The success of DD stems from the design of its loss function consisting of both the fitting term and the smoothing term. As we mentioned in Section III.B, the fitting term acts to move latent features away from each other. Therefore, using only the fitting term ignores the manifold structure and would result in a latent feature space similar to PID. The smoothing term helps DD consider the manifold structure by constraining neighboring latent features to have similar ranking scores. Combining the fitting term and the smoothing term (with proper $k$) enables DD to transform the latent features while preserving the proximity of data points in neighborhoods on the same submanifold.

Fig. 8 also shows that using an inappropriate value for $k$ leads DD to training failure. When $k$ is small, i.e., $k$=5, the latent features after the training of 6,000 epochs are split into numerous small clusters and retrieval accuracy scarcely changes from the initial MAP score. Or, if $k$ is too large, i.e., $k$=200, a collapse of the latent feature space occurs. The collapse occurs since the smoothing term acts strongly at a large scale. Each latent feature attracts 200 of its neighbors out of 1,000 training samples. Such a global influence force all the features to converge at the same point in the latent space. As also demonstrated in Fig. 3, the smoothing term and its hyperparameter $k$ have a significant impact on the retrieval accuracy of DD.

## VI. CONCLUSION AND FUTURE WORK

This paper tackled the problem of unsupervised learning of feature representations suitable for multimedia information *retrieval*. To obtain retrieval-adapted features without relying on semantic labels, we proposed the novel algorithm called DeepDiffusion (DD). The DD exploits diffusion distances on a latent feature manifold to optimize a feature extraction and embedding as well as a distance metric among embedded features. The optimization is achieved by using the carefully designed loss function, named Latent Manifold Ranking (LMR) loss, which encourages to form the latent feature space suitable for computing similarities among data samples. The comprehensive evaluation showed the following advantages and limitation of the DD algorithm.

- DD is capable of learning features more adapted to information retrieval than the existing deep learning-based unsupervised feature learning algorithms.

- DD is versatile as it learns retrieval-adapted features regardless of the data types, the datasets, and the encoder DNN architectures we have tried.

- DD is reasonably efficient on datasets having less than 100K training samples. However, for larger

datasets, DD suffers from high spatial and temporal computational costs.

We also evaluated a difference in training behavior among the DD algorithm and the existing unsupervised feature learning algorithms. The visualization of the latent feature space demonstrated that DD with an appropriate hyperparameter learns a distance metric that reflects the manifold structure in an initial latent feature space.

Future work includes evaluating the DD algorithm on datasets having more samples (e.g., 1M) and datasets consisting of multimedia data other than 3D shape and 2D image (e.g., text document). As we discussed in the section on experiments, the current DD algorithm has difficulty in training using very large datasets. We will thus consider improving computational efficiency of the DD. Also, we intend to extend the DD algorithm to a scenario of semi-supervised or supervised learning of retrieval-adapted feature representations.